\documentclass{article}
\usepackage{hyperref}
\usepackage{spconf}
\ninept

\usepackage{graphicx}
\usepackage{amsthm}
\usepackage{amsmath}
\usepackage{amssymb}
\usepackage{color}
\usepackage{parskip}
\usepackage{mwe}
\usepackage[utf8]{inputenc}

\newtheorem{definition}{Definition}

\usepackage{tikz}
\usepackage{pgfplotstable}
\usepackage{makecell}
\pgfplotsset{compat=1.14}
\usepgfplotslibrary{statistics}
\usepgfplotslibrary{groupplots}

\title{Introducing Graph Smoothness Loss for Training\\Deep Learning Architectures}

\name{\normalsize{Myriam Bontonou$^{\star,\dagger,\diamond}$ , Carlos Lassance$^{\star,\dagger,\diamond}$ , Ghouthi Boukli Hacene$^{\dagger,\diamond}$ , Vincent Gripon$^{\dagger,\diamond}$, Jian Tang$^{\circ}$, Antonio Ortega$^{\wedge}$}
\thanks{This work was supported in part by the Brittany region. Computations were performed using a Titan-V, courtesy of NVIDIA. $^\star$ Authors with equal contribution}}

\address{$^{\dagger}$ Université de Montréal, Mila\\ $^{\diamond}$ IMT Atlantique, Lab-STICC \\ $^{\circ}$ HEC Montréal, Mila \\ $^{\wedge}$ University of Southern California\\}

\begin{document}

\maketitle
\begin{abstract}
We introduce a novel loss function for training deep learning architectures to perform classification.
It consists in minimizing the smoothness of label signals on similarity graphs built at the output of the architecture. 
Equivalently, it can be seen as maximizing the distances between the network function images of training inputs from distinct classes.
As such, only distances between pairs of examples in distinct classes are taken into account in the process, and the training does not prevent inputs from the same class to be mapped to distant locations in the output domain.
We show that this loss leads to similar performance in classification as architectures trained using the classical cross-entropy, while offering interesting degrees of freedom and properties. We also demonstrate the interest of the proposed loss to increase robustness of trained architectures to deviations of the inputs.
\end{abstract}

\section{Introduction}   
 
 In machine learning, classification is one of the most studied problems. It consists in finding a function that associates inputs (typically tensors) with labels in a finite alphabet, by training on a finite number of examples. When a lot of training data is available, deep learning networks are a standard solution. A deep learning architecture is an assembly of trainable linear functions composed with nontrainable nonlinear functions. It is typically trained using variants of the Stochastic Gradient Descent (SGD) algorithm. The objective is to minimize a loss function, measuring the gap between the output of the network and the provided expected output.
 
 Cross-entropy is the most popular loss function for computer vision tasks. 
 It is often preferred over mean squared error because it converges faster and tends  to reach better accuracy.
 However, cross-entropy requires the outputs of the network to be one-hot-bit encoded vectors of the classes. This comes with noticeable drawbacks:
 \begin{itemize}
     \item The dimension of the output vectors has to be equal to the number of classes, preventing an easy adaptation to the introduction of new classes. In scenarios where the number of classes is large, this also causes the last layer of the network to contain a lot of parameters.
     \item Inputs of the same class are forced to be mapped to the same output, even if they belong to distinct clusters in the input space. This might cause severe distortions in the topological space that are likely to create vulnerabilities to small deviations of the inputs.
     \item The arbitrary choice of the one-hot-bit encoding is independent of the distribution of the input and of the initialization of the network parameters, which can slow and harden the training process.
 \end{itemize}

 To overcome these drawbacks, authors have proposed several solutions. In~\cite{hermans2017defense}, the authors propose to train using triplets, where the first element is the example to train, the second belongs to the same class and the last to another class. The objective is to minimize the distance to the element of the same class while maximizing the distance to the one of the other class. In~\cite{hinton2015distilling}, the authors replace one-hot-bit encoded vectors with soft decisions given by a pre-trained classifier. In~\cite{szegedy2016rethinking}, the authors propose to smooth the outputs of the training set. In~\cite{dietterich1994solving}, the authors propose to use error correcting codes to generate outputs of the network.
 
 In this paper, we tackle the problem of training deep learning architectures for classification without relying on arbitrary choices for the representation of the output. We introduce a loss function that aims at maximizing the distances between outputs of different classes. It is expressed using the smoothness of a label signal on similarity graphs built at the output of the network. The proposed criterion does not force the output dimension to match the number of classes, can result in distinct clusters in the output domain for a same class, and builds upon the distribution of the inputs and the initialization of the network parameters.
 We demonstrate the ability of the proposed loss function to train networks with state-of-the-art accuracy on common computer vision benchmarks and its ability to yield increased robustness to deviations of the inputs.

 The outline is as follows: in Section~\ref{rw}, we present related work. In Section~\ref{methodo}, we define formally the proposed loss and discuss its properties. In Section~\ref{experiments}, we derive and discuss experiments. Section~\ref{conclusion} is a conclusion.

\section{Related work}
 \label{rw}
 We propose to train deep learning architectures for classification, by learning an embedding of the inputs that maps examples corresponding to distinct classes to points away from each other in the embedded space. This idea can be linked to metric learning methods for $k$-nearest neighbors classification~\cite{weinberger2009distance, davis2007information}.
 
 Such objectives are particularly interesting for some types of classifications. An example is the problem of person re-identification, where deep metric learning methods have been proposed~\cite{yi2014deep, chopra2005learning}. One of them, that is known to obtain state-of-art performance, is called triplet loss~\cite{hoffer2015deep, schroff2015facenet,hermans2017defense}. In triplet loss, the idea is to enforce the distances between outputs corresponding to elements of a same class to be smaller than the ones between elements of distinct classes. This is similar to the idea of the loss introduced in this paper. 
 The main difference is that while we aim to maximize distances between elements of distinct classes, we do not enforce any constraints on the distances between elements in a same class. In a similar fashion, in~\cite{pmlr-v2-salakhutdinov07a} the authors propose the soft nearest neighbor loss, which measures the entanglement over the labeled data, and has been recently used as a regularizer in~\cite{frosst2019analyzing}. In~\cite{svoboda2018peernets}, the authors define a peer regularization layer, where latent features are conditioned on the structure induced by the graph.
 
 Other approaches propose to replace the one-hot-bit encoding of the outputs with soft values. For example in~\cite{hinton2015distilling}, the authors use the outputs of a pre-trained big network as ``soft labels'' to train a smaller one. In~\cite{szegedy2016rethinking, pereyra2017regularizing}, the outputs are smoothed to ease the training process. Other works propose to replace the classical one-hot-bit encoding of the outputs with binary codewords of an error correcting code~\cite{yang2015deep, dietterich1994solving}, thus enlarging the distance between ``class vectors''. 
 
 Our proposed method uses the notion of graph signal smoothness defined in the domain of Graph Signal Processing~\cite{shuman2013emerging,ortega2018graph}. There has been a growing interest to apply graph theories to deep neural networks, for example to interpret them~\cite{anirudhmargin}, to study their robustness~\cite{lassance2018laplacian}, or to measure the separation of classes in intermediate representations of the network~\cite{gripon2018inside}. In particular, in~\cite{gripon2018inside} the authors suggest that graph smoothness is a good measure of class separation.  

\section{Methodology}
 \label{methodo}
 
\subsection{Basic concepts}
 
 Consider a classification function $f$ that we aim at training using a dataset $X_{train} = \{(\mathbf{x}_\mu, \mathbf{y}_\mu)\}$ made of $n$ elements, where $\mathbf{x}_\mu$ refers to an input tensor and $\mathbf{y}_\mu$ to the corresponding output vector. We denote $C$ the number of classes.
 
 In the context of deep learning, $\mathbf{y}_\mu$ is typically a one-hot-bit encoded vector of its class ($\mathbf{y}_\mu \in \mathbb{R}^C$) and the network function $f$ is trained to minimize the {\em cross-entropy loss} defined as:
 
 \begin{equation}
    \mathcal{L}_{ce}(f) = -\sum_{\mu} \mathbf{y}_\mu \log(f(\mathbf{x}_\mu))\;.
    \label{eq:ce}
 \end{equation}
 
In this paper, we will consider graphs, defined $G = \langle V, \mathbf{W}\rangle$, where $V$ is the finite set of vertices and $\mathbf{W}$ is the weighted adjacency matrix: $\mathbf{W}[\mu\nu]$ is the weight of the edge between vertices $\mu$ and $\nu$, or 0 if no such edge exists. The (combinatorial) Laplacian of a graph $G = \langle V, \mathbf{W}\rangle$ is the operator $\mathbf{L} = \mathbf{D} - \mathbf{W}$ where  $\mathbf{D}$ is the degree matrix of the graph defined as:
 \[
     \mathbf{D}[\mu\nu] = \left\{ \begin{array}{cl}\displaystyle{\sum_{k \in V}{\mathbf{W}[\mu k]}} & \text{if } \mu = \nu\\ 0 & \text{otherwise}\end{array}\right.\;.
 \]
 Given a graph $G = \langle V, \mathbf{W}\rangle$ and a vector $\mathbf{s}\in \mathbb{R}^{V}$, referred to as a signal in the remaining of this work, we define the {\em graph signal smoothness}~\cite{shuman2013emerging} of $\mathbf{s}$ as:
 
 \begin{eqnarray*}
     s_{G}(\mathbf{s}) &=& \mathbf{s}^\top \mathbf{L} \mathbf{s} \\&=& \sum_{\mu, \nu \in V}{\mathbf{W}[\mu\nu]\left(\mathbf{s}[\mu] - \mathbf{s}[\nu]\right)^2}\;,
 \end{eqnarray*}
 where $\mathbf{}^\top$ is the transpose of $\mathbf{s}$.
 
 Finally, we call \emph{label signal} associated with the class $c$ the binary indicator vector $\mathbf{s}_c$ of elements of class $c$. Hence, $\mathbf{s}_c[\mu] = 1$ if and only if $\mathbf{x}_\mu$ is in class $c$.
 
 \subsection{Proposed graph smoothness loss}
 We propose to replace the cross-entropy loss with a graph smoothness loss.
 Consider a fixed metric $\|\cdot\|$. We compute the distances between the representations $f(\mathbf{x}_\mu), \forall \mu$. Using these distances, we build a $k$-nearest neighbor graph $G^{nn}_k = \langle V_k, \mathbf{W}_k\rangle$ containing $n$ vertices. We apply a kernel parameterized by $\alpha$ to obtain each element of $\mathbf{W}$: $$\mathbf{W}_k[\mu\nu] \neq 0 \Rightarrow \mathbf{W}_k[\mu\nu] = \exp{\left(-\alpha\|f(\mathbf{x}_\mu) - f(\mathbf{x}_\nu)\|\right)},\forall \mu, \forall \nu\;.$$
 
 We call the resulting graph $G_k = \langle V_k, \mathbf{W}_k\rangle$ the similarity graph of $f$ of parameter $k$.
 
 \begin{definition}
    We call $\emph{graph smoothness loss}$ of $f$ the quantity:
    \begin{eqnarray*}
        \mathcal{L}_{G_k}(f) &=& \sum_{c=1}^{C}{s_{G_k}(\mathbf{s}_c)}\\
        &=& \hspace{-1.2cm}\underbrace{\sum_{\mathbf{x}_\mu, \mathbf{x}_{\nu}, \mathbf{W}_k[\mu \nu] \neq 0\atop \mathbf{s}_c[\mu] \mathbf{s}_{c}[\nu] = 0, \forall c}}_{\text{sum over inputs of distinct classes}}{\hspace{-1cm}\exp{\left(-\alpha\|f(\mathbf{x}_{\mu}) - f(\mathbf{x}_{\nu})\|\right)}}
        \;.
    \end{eqnarray*}
 \end{definition}
 
 In the following subsection, we motivate the use of this loss.
 
\subsection{Properties}

 \begin{figure*}[h!]
 \hspace{-1cm}\includegraphics[scale=0.45]{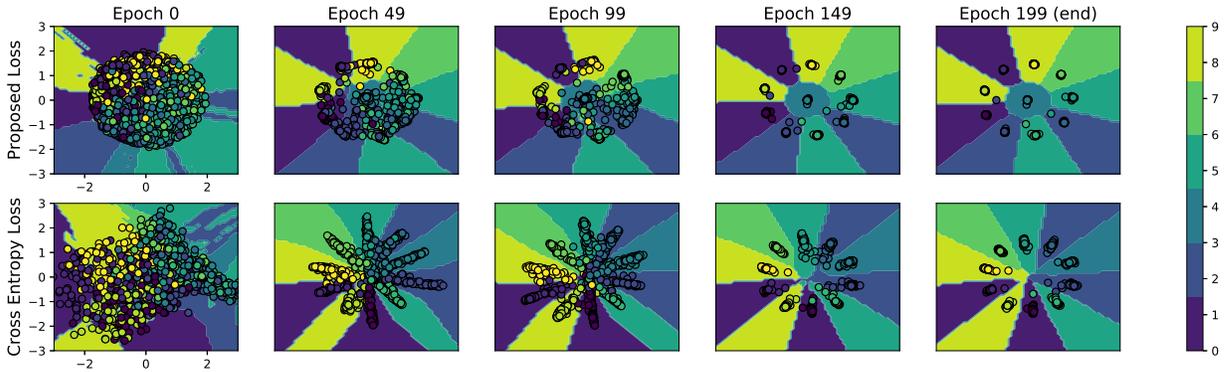}
 \caption{\label{Exp1}Embeddings of CIFAR-10 training set learned using the proposed graph smoothness loss with $d=2$ (top row) compared with the ones obtained using a bottleneck layer and cross-entropy with the same architecture (bottom row).}
 \end{figure*}
 
 The cross-entropy loss introduced in~\eqref{eq:ce} aims at mapping inputs of the network to arbitrarily chosen one-hot-bit encoded vectors representing the corresponding classes. Our proposed loss function differs from the cross-entropy loss in several ways:
 \begin{itemize}
     \item The cross-entropy loss forces a mapping from the input to a single point for each class. This might force the network to considerably distort space, for example in the case where a class is made of several disjoint clusters. The use of $k$-nearest neighbors gives more flexibility to the proposed loss: using a small value of $k$ it is possible to minimize the graph smoothness loss with multiple clusters of points for each class.
     \item The cross-entropy loss requires to arbitrarily choose the outputs of the network, disregarding the dataset and the initialization of the network. In contrast, the proposed loss is only interested in relative positioning of outputs with regards to one another, and can therefore build upon the initial distribution yielded by the network.
     \item The cross-entropy loss forces to use an output vector whose dimension is the number of classes of the problem at hand. It is thus required to modify the network to accommodate for new classes (e.g. in an incremental scenario). The dimension of the network output $d$ is less tightly tied to the number of classes with the proposed classifier.
 \end{itemize}
 
 It is important to note that the capacity and the dimension of the output space should be bound to the problem to be solved. On the one hand, if the dimension of the output domain is too low, it is likely that the network will fail to converge: consider a toy example where we try to separate $n$ data points so that they are all at the same distance in the output space. This is only possible if the dimension of the output space is at least $n-1$. On the other hand, if the capacity of the output space is too large, a trivial solution to minimize the loss consists in arbitrarily scattering the inputs, so that the distance between the image of any two inputs becomes large. This relation between the dimension of the output space and the ability of the network to classify is further discussed in the next section.
 
\section{Experiments}
\label{experiments}

\subsection{Experimental set-up}
 We evaluate the performance of the proposed loss using three common datasets of images, namely CIFAR-10/CIFAR-100~\cite{krizhevsky2009learning} and SVHN~\cite{netzer2011reading}. For each dataset, we follow the same experimental process: a) We pick an optimized network architecture for the dataset; b) We build a network with the same architecture (number of layers, number of features per layer) and hyperparameters (number of epochs, learning rate, gradient descent algorithm,  mini-batch size,  weight decay, weight normalization), but we replace the cross-entropy loss with the proposed graph smoothness loss; c) We tune the loss additional hyperparameters ($k$, $\alpha$, $d$). When performing classification, we add a simple classifier on top of the network to measure its accuracy.
 All networks are using PyTorch~\cite{paszke2017automatic}. Note that all input images are normalized before being processed.
 It is important to keep in mind that by choosing this methodology, we bias the experiments in favor of using the cross-entropy loss, since the chosen architectures have been designed for its use.
 
 The network architecture we use for CIFAR-10, CIFAR-100 and SVHN is PreActResNet-18~\cite{he2016identity}, as implemented in~\cite{zhang2017mixup}. The network is trained for 200 epochs using 100 examples per mini-batch. SVHN and CIFAR-10 networks are trained with SGD, using a learning rate that starts at 0.1 and is divided by 10 at epochs 100 and 150, with a weight decay factor of $10^{-4}$ and a Nesterov momentum~\cite{pmlr-v28-sutskever13} of 0.9. On the other hand, CIFAR-100 is trained with the Adam optimizer~\cite{kingma2014adam}, using a learning rate that starts at 0.001 and is divided by 10 at epochs 100 and 150.  Note that a graph is built for each mini-batch (graph smoothness is calculated on a graph of 100 vertices). We do not use dropout or early stopping in any of our experiments.
 
 In the two above networks, the linear function of the last layer outputs a $C$ dimensional vector on which a softmax function is applied. When using the proposed loss, the linear function outputs a $d$ dimensional vector ($d$ is an hyperparameter), normalized with respect to the $\mathcal{L}_2$ norm. We use this normalization to constrain the outputs to remain in a compact subset of the output space. As discussed in Section~\ref{methodo}, if we did not, and since we use the $\mathcal{L}_2$ metric to build the graphs in our experiments, the network would likely converge to a trivial solution that would scatter the outputs far away from each other in the output domain, regardless of their class.
 
\subsection{Visualization}
We first compare the embedding obtained using the proposed loss and $d=2$ with the one obtained when putting a bottleneck layer of the same dimension $d=2$ using the cross-entropy loss. Results on CIFAR-10 are depicted in Figure~\ref{Exp1}.
Exceptionally, for this experiment we do normalize the output of the last layer of the network using batch norm instead of $\mathcal{L}_2$ norm. This is because using $d=2$ with a $\mathcal{L}_2$ normalization would reduce the output space dimension to 1, which would likely be too small to allow the training loss to descend to 0. We observe that in the third column of Figure~\ref{Exp1}, our method creates clusters whereas the baseline method creates lines. This reflects the choice of the distance metric: our method uses the $\mathcal{L}_2$ distance, whereas the baseline seems to use the cosine distance instead. Figure~\ref{Exp1} shows that training examples are better clustered at the end of the training process when using the proposed loss than with the cross-entropy loss.

\subsection{Classification}
\begin{figure}[ht]
 \begin{center}
   \begin{tikzpicture}[]
       \begin{axis}[
           xlabel=$\alpha$,
           ylabel=Test accuracy,
           ymin=0,
           ymax=100,
           xmode = log,
           legend pos=south east, width=8cm,height=4cm]

         \addplot table {cifar_10/testsetXalpha.txt};
         \addlegendentry{CIFAR-10}
         \addplot table {cifar_100/testsetXalpha.txt};
         \addlegendentry{CIFAR-100}
         \addplot [mark=triangle] table {svhn/testsetXalpha.txt};
         \addlegendentry{SVHN}
       \end{axis}
   \end{tikzpicture}
    \caption{\label{Fig:testxalpha} Test set accuracy as a function of $\alpha$.} 
 \end{center}
\end{figure}
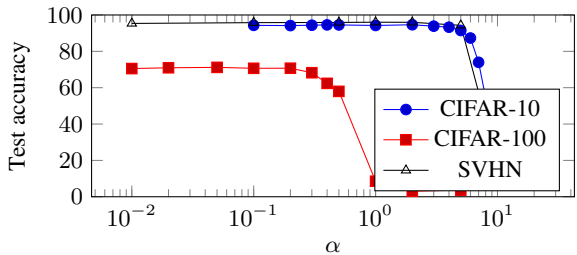

We evaluate the influence on classification performance of the three hyperparameters of the proposed loss: the number of neighbors $k$ to consider in the similarity graph $G_k$, the number of dimensions $d$ coming out of the network and the scaling parameter $\alpha$ used to define the weights of the graph. When varying $k$, we fix $d$ to be the number of classes and $\alpha = 2$; when varying $d$, we fix $k$ to the maximum value and $\alpha = 2$ and when varying $\alpha$, we fix $d$ to be the number of classes and $k$ to the maximum value. The results are summarized in Figures~\ref{Fig:testxk},~\ref{Fig:testxd},~\ref{Fig:testxalpha}. Note that a 10-NN classifier was used to obtain the accuracy. We observe that the higher $k$ is, the higher the test accuracy is, even if the sensitivity to $k$ is lower when $k$ is larger than the number of classes. As soon as $d$ becomes large enough to accommodate for the number of classes, we observe that the test accuracy starts dropping slowly. Therefore, because using a larger value of $d$ does not seem particularly harmful, applications where the number of classes is unknown (such as in incremental learning) should use a high $d$. Similarly, there is almost no dependence to $\alpha$ as long as its value is small enough. Indeed, when $\alpha$ is large, the loss tends to be close to 0 even if the corresponding distances are still relatively small.

\begin{figure}[ht]
 \begin{center}
   \begin{tikzpicture}[]
       \begin{axis}[
           xlabel=$k$,
           ylabel=Test accuracy,
           ymin=0,
           ymax=100,
           xmode = log,
           legend pos=south east, width=8cm,height=4cm]
         \addplot table {cifar_10/testsetXk.txt};
         \addlegendentry{CIFAR-10}
         \addplot [mark=triangle] table {svhn/testsetXk.txt};
         \addlegendentry{SVHN}
       \end{axis}
   \end{tikzpicture}
    \caption{\label{Fig:testxk} Test set accuracy as a function of $k$.}
 \end{center}
\end{figure}
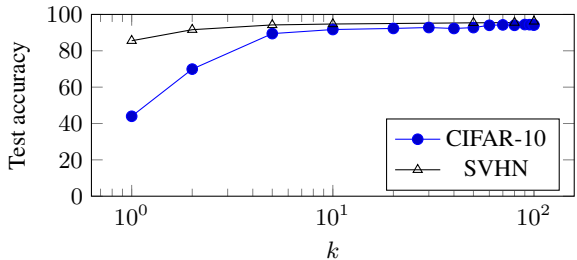

\begin{figure}[ht]
 \begin{center}
   \begin{tikzpicture}[]
       \begin{axis}[
           xlabel=$d$,
           ylabel=Test accuracy,
           ymin=0,
           ymax=100,
           xmode=log,
           legend pos=south east,
           width=8cm,height=4cm]
         \addplot table {cifar_10/testsetXd.txt};
         \addlegendentry{CIFAR-10}
         \addplot table {cifar_100/testsetXd.txt};
         \addlegendentry{CIFAR-100}
         \addplot [mark=triangle] table {svhn/testsetXd.txt};
         \addlegendentry{SVHN}
       \end{axis}
   \end{tikzpicture}
    \caption{\label{Fig:testxd} Test set accuracy as a function of $d$.}
 \end{center}
\end{figure}
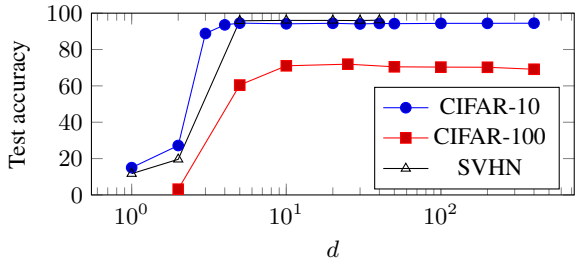

We next evaluate the performance of the graph smoothness loss for classification. To this end, we compare its accuracy to that achieved with optimized network architectures using a cross-entropy loss (CE). We use various classifiers on top of the graph smoothness loss-trained architectures: a $1$-nearest neighbors classifier ($1$-NN), a $10$-nearest neighbors classifier ($10$-NN) and a support vector classifier (SVC) using radial basis functions. The results are summarized in Table~\ref{table:performance}. We observe that the test error obtained with the proposed loss is close to the CE test error, suggesting that the proposed loss is able to compete in terms of accuracy with the cross-entropy. Interestingly, we do not observe a significant difference in accuracy between the classifiers. Besides, both losses require the same training time.

\begin{table}[ht]
\begin{center}
\begin{tabular}{c|ccc}
\hline
Loss - Classifier          & CIFAR-10 & CIFAR-100 & SVHN \\
     \Xhline{2\arrayrulewidth}
CE - Argmax                & \textbf{5.06\%}   & \textbf{27.92\%}   &  3.69\%        \\
Proposed      - 1-NN       & 5.63\%   & 29.17\%   &  3.84\%         \\
Proposed      - 10-NN      & 5.48\%   & 28.82\%   &  \textbf{3.34}\%            \\
Proposed      - RBF SVC    & 5.50\%   & 30.55\%   &    3.40\%   \\ \hline 

\end{tabular}
\end{center}
\caption{Test errors on CIFAR-10, CIFAR-100 and SVHN datasets. The top contains the test error of the optimized network architectures for a cross-entropy loss (CE). The bottom contains the test error of the same network architectures for our proposed graph smoothness loss, associated with three different classifiers.}
\label{table:performance}
\end{table}

\subsection{Robustness}

We evaluate the robustness of the trained architectures to deviation of inputs in Table~\ref{table:robustness}. We first report the error rate on the clean test set for which we observe a small drop in performance when using the proposed loss. However, this drop is compensated by a better accommodation to deviations of the inputs, as reported by the Mean Corruption Error (MCE) scores (see~\cite{hendrycks2019robustness}). Such a trade-off between accuracy and robustness has been discussed in~\cite{Fawzi2018}. For this experiment, we fixed $k$ to its maximum value, $d=200$, $\alpha=2$ and we used 10-NN as a classifier when using the graph smoothness loss.

\begin{table}[ht]
\centering
\begin{tabular}{lc|cc}
\hline
Method               & Clean test error & MCE   & relative MCE \\         \Xhline{2\arrayrulewidth}
Cross-entropy              & {\bf 5.06}\%           & 100   & 100          \\
Proposed             & 5.60\%           & {\bf 95.28} & {\bf 90.33}  \\ \hline     
\end{tabular}
\caption{Robustness comparison on the 15 corruptions benchmarks from~\cite{hendrycks2019robustness} on the CIFAR-10 dataset.}
\label{table:robustness}
\end{table}

\section{Conclusion}
\label{conclusion}

In this work, we introduced a loss function that consists in minimizing the graph smoothness of label signals on similarity graphs built at the output of a deep learning architecture.
We discussed several interesting properties of this loss when compared to using the classical cross-entropy.
Using experiments, we showed the proposed loss can reach similar performance as cross-entropy, while providing more degrees of freedom and increased robustness to deviations of the inputs.

\newpage

\bibliography{main}
\bibliographystyle{IEEEbib}

\end{document}